\documentclass[runningheads, envcountsame, a4paper, hyphens, article]{llncs}
\usepackage{graphicx}
\usepackage[utf8]{inputenc}
\usepackage{amsfonts, amssymb, amsmath}
\usepackage{dsfont}
\usepackage{multirow}
\usepackage{array}
\usepackage{microtype}
\usepackage{hyperref}
\usepackage[rgb]{xcolor}
\usepackage{eucal}

\usepackage{tikz,pgf,pgfplots,pgfplotstable}
\usetikzlibrary{plotmarks}
\usetikzlibrary{shapes,arrows}
\usepackage[misc]{ifsym}

\begin{document}
\title{Deep Autoencoders for Anomaly Detection in Textured Images using CW-SSIM}
\titlerunning{Deep Autoencoders for Anomaly Detection in Textured Images}
%
\author{Andrea Bionda \and
Luca Frittoli\textsuperscript{(\Letter)} \and
Giacomo Boracchi}
\authorrunning{A. Bionda et al.}
%
\institute{DEIB, Politecnico di Milano, via Ponzio 34/5, Milan, Italy\\
\email{andrea.bionda@mail.polimi.it\\$\{$luca.frittoli, giacomo.boracchi$\}$@polimi.it}
}
\maketitle              
\begin{abstract}
Detecting anomalous regions in images is a frequently encountered problem in industrial monitoring. A relevant example is the analysis of tissues and other products that in normal conditions conform to a specific texture, while defects introduce changes in the normal pattern. We address the anomaly detection problem by training a deep autoencoder, and we show that adopting a loss function based on Complex Wavelet Structural Similarity (CW-SSIM) yields superior detection performance on this type of images compared to traditional autoencoder loss functions. Our experiments on well-known anomaly detection benchmarks show that a simple model trained with this loss function can achieve comparable or superior performance to state-of-the-art methods leveraging deeper, larger and more computationally demanding neural networks.\footnote{Final authenticated version: \url{https://doi.org/10.1007/978-3-031-06430-2_56}}

\end{abstract}

\section{Introduction}\label{sec:intro}
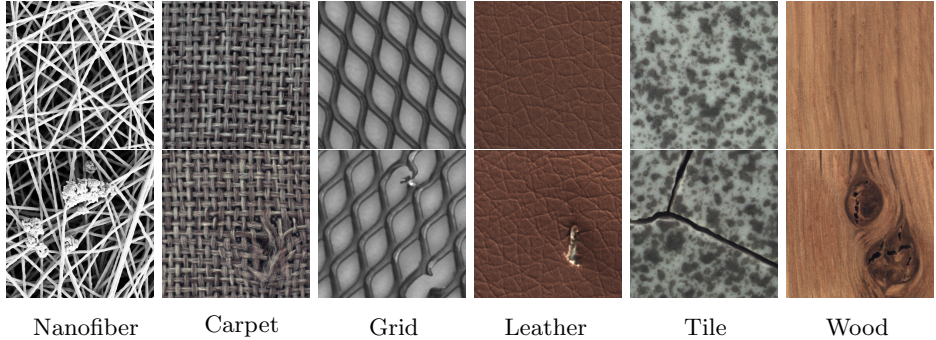
\begin{figure}
  \centering
  \begin{minipage}[b]{0.159\textwidth}
    \includegraphics[width=\textwidth]{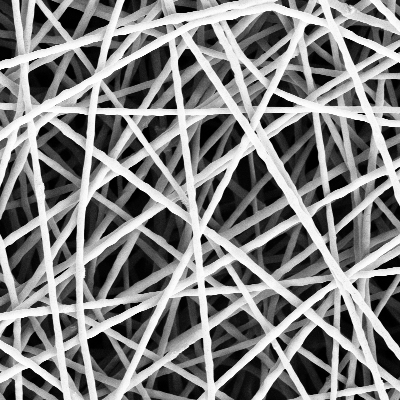}
    \includegraphics[width=\textwidth]{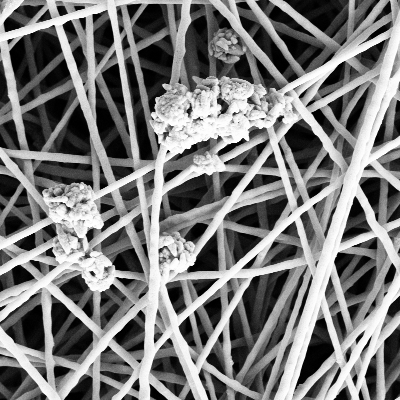}
  \end{minipage}
  \hfill
  \begin{minipage}[b]{0.159\textwidth}
    \includegraphics[width=\textwidth]{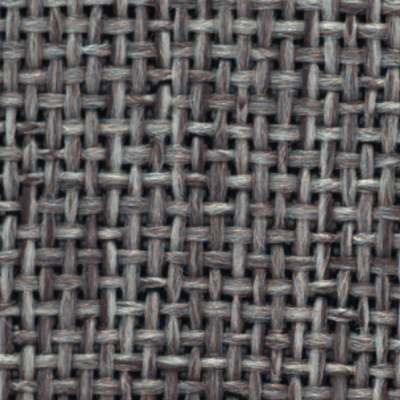}
    \includegraphics[width=\textwidth]{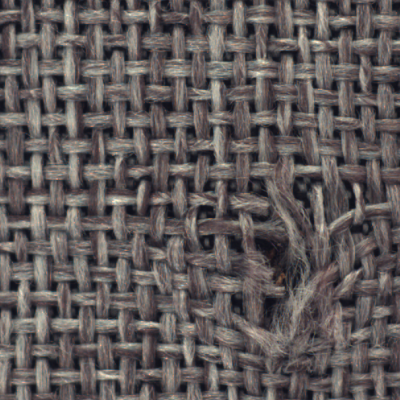}
  \end{minipage}
  \hfill
  \begin{minipage}[b]{0.159\textwidth}
    \includegraphics[width=\textwidth]{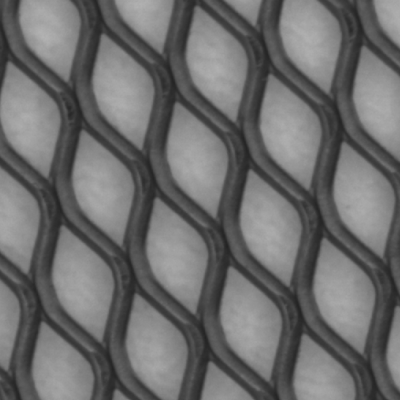}
    \includegraphics[width=\textwidth]{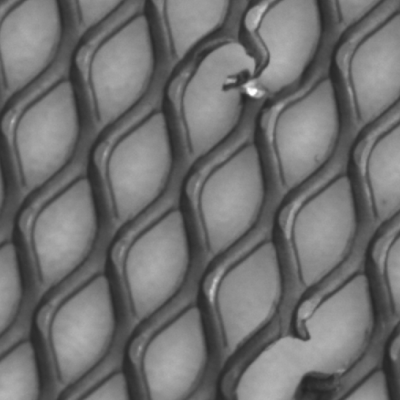}
  \end{minipage}
  \hfill
  \begin{minipage}[b]{0.159\textwidth}
    \includegraphics[width=\textwidth]{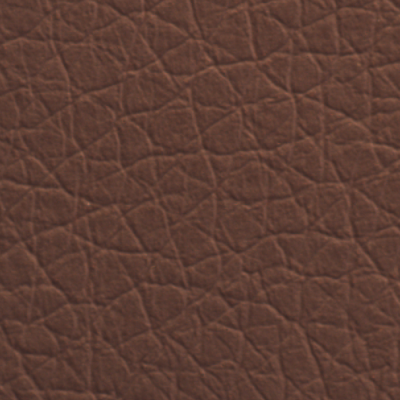}
    \includegraphics[width=\textwidth]{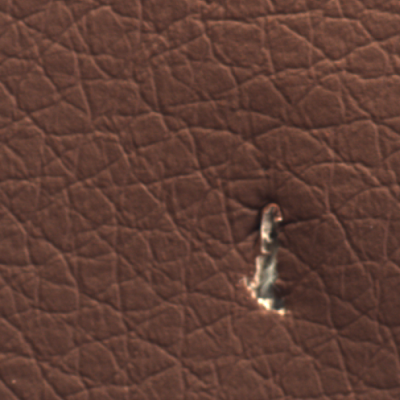}
  \end{minipage}
  \hfill
  \begin{minipage}[b]{0.159\textwidth}
    \includegraphics[width=\textwidth]{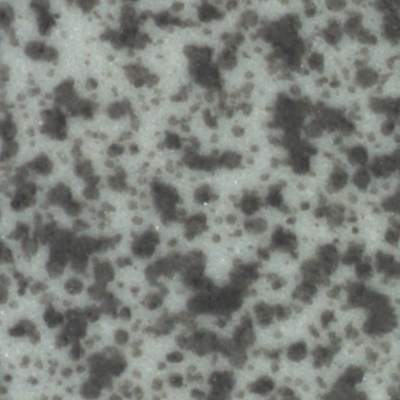}
    \includegraphics[width=\textwidth]{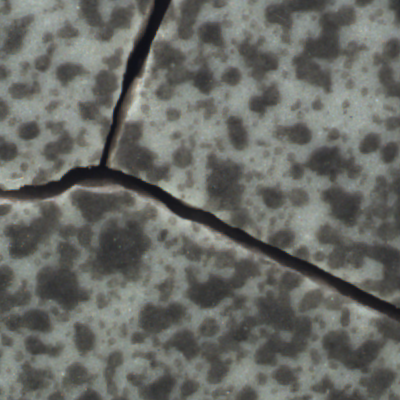}
  \end{minipage}
  \hfill
  \begin{minipage}[b]{0.159\textwidth}
    \includegraphics[width=\textwidth]{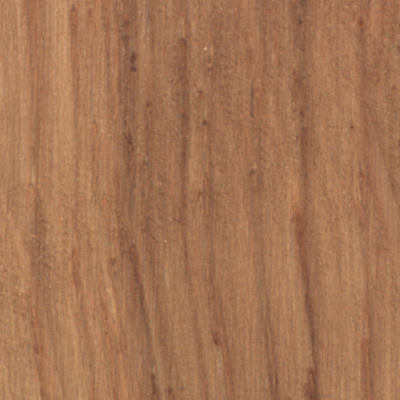}
    \includegraphics[width=\textwidth]{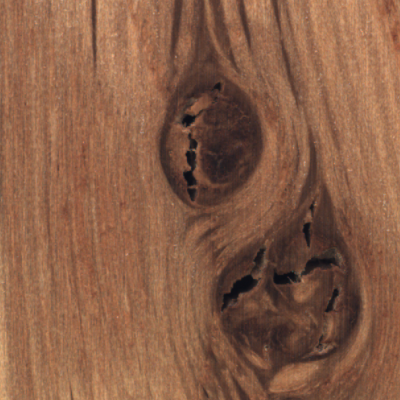}
  \end{minipage}
  \begin{tikzpicture}[overlay]
	    \node at (-5.05,-0.0) {\small Nanofiber};
	    \node at (-3,-0.) {\small Carpet};
	    \node at (-1,-0.) {\small Grid};
	    \node at (1,-0.) {\small Leather};
	    \node at (3.1,-0.) {\small Tile};
	    \node at (5.1,-0.) {\small Wood};
  \end{tikzpicture}
  \caption{Examples of normal (above) and anomalous images (below) from the Nanofiber~\cite{carrera2016defect} and MVTec~\cite{bergmann2021mvtec} datasets.
  }\label{fig:Nanofiber_good}
  \vspace{-5mm}
\end{figure}

The automatic detection and localization of anomalous regions within images is a crucial problem in many industrial scenarios where the production volume prevents human supervision. In most cases, anomaly detection solutions are required to identify regions that deviate either in appearance, structure or contrast from the rest of the image, which instead follows a pattern or texture representing normal conditions. As an example, consider images of nanofiber tissues~\cite{bolgen2016nanofibers}, namely polymer fibers with diameters down to one hundred nanometers, which look like a texture of randomly overlapping filaments. In this context, it is crucial to automatically locate defects, i.e., regions that deviate from this texture, a problem that was first addressed in~\cite{carrera2016defect}. Figure \ref{fig:Nanofiber_good} represents the textures from the Nanofiber~\cite{carrera2016defect} and MVTec dataset~\cite{bergmann2021mvtec}, and different kinds of anomalies. 

Nanofiber images are a relevant example of many industrial monitoring scenarios where normal images are characterized by specific textures or repetitive patterns that can differ in orientation, scale and shape from image to image. For this reason, it is rarely possible to detect anomalies in test images by direct comparison against a template or a reference. Moreover, anomaly detection is performed by assigning an anomaly score to each pixel of the image, while the vast majority of anomaly detection methods is designed to classify an entire image as normal or anomalous~\cite{ruff2021unifying}. In our case, anomaly detection is seen as a segmentation problem where images are to be segmented into normal and anomalous regions. Since anomalies are typically rare, and their annotation is expensive and time-consuming, supervised learning is impractical in most industrial scenarios. Thus, anomaly detection is usually addressed by unsupervised models trained exclusively on normal images, which are usually abundant and do not require annotation~\cite{bergmann2021mvtec}. 

A mainstream approach to detect anomalous regions consists in processing the image in small patches either by sparse representations~\cite{carrera2016defect} or deep neural networks~\cite{bergmann2020uninformed,napoletano2018anomaly,zhou2017anomaly}. Here, the patch size determines a trade-off between the resolution of the predictions, which increases when using smaller patches, and the detection accuracy, which improves when using larger patches since each patch contains more information. In this respect, deep neural networks have the advantage that they can adjust their \emph{effective receptive fields} during training, and learn what is the most informative area inside a fixed-size patch. Thus, the patch size is perhaps a less influential parameter in deep neural networks than in traditional machine learning methods for anomaly detection.

\emph{Convolutional autoencoders}, namely convolutional neural networks (CNNs) trained to extract a latent representation and then reconstruct the original image, have been widely employed for anomaly detection purposes~\cite{zhou2017anomaly,bergmann2018improving}. The rationale is that an autoencoder trained exclusively on normal images will accurately reconstruct only the normal regions of a test image. Thus, it is possible to detect as anomalous those regions of a test image that substantially differ from their reconstruction. Autoencoders are usually trained to minimize the \emph{Mean Square Error} (MSE) between the input and reconstructed images. However, the MSE is not well suited for textured images since it measures the average pixelwise difference, while repetitive patterns are substantially more relevant than individual pixel values. For this reason, a loss function based on the \emph{Structural Similarity} index (SSIM)~\cite{wang2004image} can improve the anomaly detection performance~\cite{bergmann2018improving}.

We propose to address anomaly detection in textured images using convolutional autoencoders trained and tested using similarity metrics specifically designed to compare textures. In particular, we employ \emph{Complex Wavelet Structural Similarity} (CW-SSIM)~\cite{wang2005translation} to train an autoencoder, and also to compare the original and reconstructed images at test time. Using CW-SSIM substantially improves the reconstruction quality compared to MSE, and yields better robustness to differences in scale and orientation compared to SSIM. Our experiments on the Nanofiber~\cite{carrera2016defect} and MVTec Texture~\cite{bergmann2021mvtec} datasets show that a simple autoencoder can outperform comparably more complicated and computationally expensive models such as those in \cite{napoletano2018anomaly,bergmann2020uninformed}. The code is available at \url{https://github.com/AndreaBiondaPolimi/Anomaly-Detection-Autoencoders-using-CW-SSIM}.

Our solution is currently designed to process grayscale images, since CW-SSIM extracts structural and contrast information from single-channel images. This type of images can be encountered in many circumstances, e.g. where quality control is performed by X-ray imaging or electronic microscopes~\cite{carrera2016defect}. Our solution yields excellent performance even on some MVTec datasets containing RGB images, which we convert to grayscale before feeding them to our autoencoder. However, it turns out that, in some MVTec images, anomalies only affect the color of certain regions. Thus, color information is crucial for anomaly detection, which encourages us to investigate how to better handle RGB images.

\section{Problem Formulation}\label{sec:formulation}
Let us denote by $I:\mathcal{X}\rightarrow\mathbb{N}^c$ the image to be analyzed, where $\mathcal{X}\subset\mathbb{Z}^2$ is a pixel grid and $c$ is the number of color channels, whose intensity can range from $0$ to $2^d-1$, where $d$ is the color depth. 
We primarily consider grayscale images (i.e., $c=1$), since in several industrial monitoring scenarios images are acquired by X-ray sensors or electronic microscopes~\cite{carrera2016defect}. Our goal is to locate anomalous regions in $I$, i.e., to estimate the unknown anomaly mask $\Omega:\mathcal{X}\rightarrow\{0,1\}$:
\begin{equation}\label{eq:problem_formulation_eq_intro}
    \Omega(i,j) =   \begin{cases}
                        1 & \text{if pixel $(i,j)$ falls inside an anomalous region},\\
                        0 & \text{otherwise}.
                    \end{cases}
\end{equation}
In particular, we are interested in estimating an anomaly mask $\hat\Omega: \mathcal{X} \rightarrow \{0,1\}$ that approximates $\Omega$ as well as possible. Typically, this is done by combining anomaly scores computed on image patches, which we indicate by $x$.

We assume that a training set $TR$ containing only normal (i.e., anomaly-free) images is provided, so we address anomaly detection in an unsupervised manner. These are reasonable settings in industrial monitoring, where normal images are abundant, while anomalies are rare and expensive to annotate. Moreover, anomalies can vary in shape, size and other characteristics, and an annotated training set might not encompass all the possible anomalies that will be encountered during testing. Our goal is to detect as anomalous all the regions of $I$ that do not comply with the structure of normal images provided for training.

\section{Related Work}\label{sec:related}
Anomaly detection in images is a widely studied problem. The vast majority of the existing solutions follow a one-class classification approach, where the entire image is classified as normal or anomalous. In this paper we address the more challenging problem of detecting anomalous regions within the image, thus we refer to~\cite{ruff2021unifying} for a recent survey of the one-class classification literature. 

The first solutions proposed to detect anomalous regions in images apply unsupervised learning methods, such as \emph{One-Class Support Vector Machine} (OC-SVM)~\cite{li2003improving} or \emph{Kernel Density Estimation} (KDE)~\cite{Kde}, to features extracted from image patches. The extracted features can either be hand-crafted features, such as the response of the image to Gabor filters~\cite{tsa2000automated} or regularity measures~\cite{chetverikov2002finding}, or data-driven features such as dictionaries learned to yield \emph{sparse representations}~\cite{carrera2016defect}. Another approach models the texture by an autoregressive model and detects anomalies using the residuals~\cite{haindl2007texture}. In general, data-driven models are more effective than those based on hand-crafted features, thanks to their greater flexibility~\cite{carrera2016defect}. 

Nowadays, the vast majority of anomaly detection methods leverage features extracted by \emph{Convolutional Neural Networks} (CNNs), trained to address a different classification problem. A relevant example is the work by \emph{Napoletano et al.}~\cite{napoletano2018anomaly}, where a pre-trained ResNet-18~\cite{he2016deep} is used to extract feature vectors from normal patches. Then, these features are reduced in dimension by PCA and clustered by K-Means to form a dictionary of normal features. The anomaly score on each test patch is computed as the Euclidean distance between its projected feature vector and the most similar dictionary atom. Another way to exploit pre-trained CNNs is the \emph{Student-Teacher} approach~\cite{bergmann2020uninformed}, which is based on a feature-extracting network (pre-trained on natural images), referred to as the Teacher, and a set of Student networks trained on anomaly-free patches to estimate the Teacher's output in a regression problem. During testing, the anomaly score of each pixel is computed by combining the regression error and the prediction variance of the Student networks. The intuition is that the Students can extract similar features compared to the Teacher only from normal patches.

The most popular deep learning methods for anomaly detection in images are autoencoders~\cite{zhou2017anomaly}, namely CNNs trained to extract a compact \emph{latent representation} from the input image and then to reconstruct the input image from the \emph{latent representation}. The most common loss function is simply the MSE between the input and reconstructed images, even though \emph{Bergmann et al.}~\cite{bergmann2018improving} show that using SSIM~\cite{wang2004image} as training loss instead of the MSE substantially improves the reconstruction quality, as it takes into account local structures instead of differences in individual pixel values. During testing, the autoencoder can accurately reconstruct only patches that are similar to the anomaly-free training data, thus anomalous regions can be detected by analyzing the pixelwise reconstruction error. In this work, we extend~\cite{bergmann2018improving} and demonstrate that a loss function based on CW-SSIM~\cite{wang2005translation} further improves the anomaly detection performance. 

It is also worth mentioning some more sophisticated anomaly detectors based on \emph{Generative Adversarial Networks} (GANs)~\cite{akcay2018ganomaly,schlegl2019f} and \emph{reconstruction by inpainting}~\cite{zavrtanik2021reconstruction}, which are becoming increasingly popular. However, we do not consider these methods since we primarily focus on autoencoders, and investigate the impact of using different loss functions. Moreover, so far these solutions based on GANs have been applied to produce a single anomaly score for the whole input image, and cannot be easily adapted to localize anomalies within images~\cite{ruff2021unifying}.

\section{Proposed Solution}\label{sec:proposed}
We address anomaly detection using an autoencoder~\cite{zhou2017anomaly} trained on small image patches. Autoencoders consist of two sub-networks: an \emph{encoder} $\mathcal E$ and a \emph{decoder} $\mathcal D$. The encoder extracts a feature vector $z=\mathcal E(x)$, called \emph{latent representation}, from a patch $x$ of the input image. The decoder is an upsampling network taking as input $z$ and returning a patch $y=\mathcal D(z)$ having the same size as $x$. The two sub-networks are jointly trained on patches of images from $TR$, which do not contain anomalies, to minimize a loss function measuring the pixelwise reconstruction error, namely the difference between $y$ and $x$. This loss is typically defined as the MSE, but in~\cite{bergmann2018improving} it has been shown that replacing the MSE with structural similarity metrics such as SSIM~\cite{wang2004image} improves the reconstruction and anomaly detection performance. 

Structural similarity metrics were initially designed to classify textured images~\cite{zujovic2009structural} and to assess their quality~\cite{wang2004image} by detecting differences with respect to a template. We focus on metrics based on the \emph{steerable filter decomposition}~\cite{Steerable_filters}, which can extract informative features describing a texture. For this reason, we propose to train the autoencoder to maximize the Complex Wavelet Structural Similarity (CW-SSIM) index~\cite{wang2005translation} between input and reconstructed images. We also employ CW-SSIM at test time to estimate the anomaly map by comparing the input and reconstructed images. Compared to other structural similarity metrics (including SSIM), CW-SSIM is more robust to scaling, translations and rotations, and we show that this further improves the reconstruction quality and the anomaly detection ability compared to the autoencoder in~\cite{bergmann2018improving}. 

In what follows we first illustrate the steerable filter decomposition (Section~\ref{subsec:Steerable}), then we introduce the CW-SSIM index and illustrate how we employ it in the loss function of the autoencoder (Section~\ref{subsec:CWSSIM}). Finally, we present our anomaly detection procedure, which is also based on CW-SSIM (Section~\ref{subsec:reco_based}).

\subsection{Steerable Filter Decomposition}\label{subsec:Steerable}
Steerable filter decomposition is based on recursive application of filtering and subsampling operations, stacked in the \emph{steerable pyramid algorithm}. It has been demonstrated that the statistics obtained by this procedure can characterize well a broad class of textures~\cite{CW_Extraction}. This algorithm decomposes the input patch $x$ into $M$ different \textit{subbands} $x^m$ in the frequency domain, where $m\in\{1,\ldots,M\}$, at $S$ different scales. The first subband is obtained by applying the \emph{Fast Fourier Transform} (FFT) to the original patch. The patch is then downsampled by $2 \times 2$ average pooling and convolved with $O$ oriented kernels $S-2$ times to obtain the subbands at different scales. The patch is downsampled one more time to obtain the last subband. In the end we have
\begin{equation}\label{eq:subbands_total}
    M=O(S-2) + 2
\end{equation}
subbands, where the term ``$+2$'' takes into account the first and the last subbands. The design of the filters and more implementation details can be found in \cite{Shiftable,Steerable_filters}. 

\subsection{CW-SSIM as Loss Function}\label{subsec:CWSSIM}
As in \cite{bergmann2018improving}, we train the autoencoder over randomly cropped patches $\{x\}$ from anomaly-free images in $TR$. During training, the loss function $\mathcal{L}$ compares each patch $x$ with the corresponding $y = \mathcal D(\mathcal E(x))$ reconstructed by the autoencoder:
\begin{equation}
    \mathcal{L}(x,y) = 1 - \frac{1}{M} \sum_{m=1}^{M} \frac{1}{L} \sum_{l=1}^{L} \text{CW-SSIM}(x^m_l,y^m_l),\label{eq:cw-ssim_pool-subband}
\end{equation}
where $x^m$ and $y^m$ indicate the subbands in which $x$ and $y$ are decomposed by the steerable filter algorithm, for $m\in\{1,\ldots,M\}$. The CW-SSIM index is computed by dividing each subband into $L$ small windows $x^m_l$ and $y^m_l$ of size $R \times R$ following a regular grid sampling with stride 1, where $l\in\{1,\ldots,L\}$. The term $\text{CW-SSIM}(x^m_l,y^m_l)$, which compares two windows at the same spatial location $l$ and subband $m$, is defined as:
\begin{equation}\label{eq:cw-ssim}
    \text{CW-SSIM}(x^m_l,y^m_l) = \frac{2 |\langle x^m_l,y^m_l \rangle| + K}
        {  \left\Vert x^m_l \right\Vert^2_2 + \left\Vert y^m_l \right\Vert^2_2 + K},
\end{equation}
where $\langle\cdot,\cdot\rangle$ and $\Vert\cdot\Vert^2_2$ indicate the inner product and norm of complex vectors, and $K$ is a small constant preventing the denominator from vanishing \cite{wang2005translation}. In \eqref{eq:cw-ssim} the windows $x^m_l,y^m_l$ are assumed to be flattened to column vectors. The loss function in \eqref{eq:cw-ssim_pool-subband} represents a dissimilarity score between $x$ and $y$ ranging in $[0, 2]$. 

\paragraph{Implementation Details.}
We empirically chose the window size $R=7$ for CW-SSIM and the number of orientations $O=6$ for the steerable filter decomposition. Since we observed that subbands with a spatial dimension lower than $16\times16$ pixels in the last subband do not contain useful information for anomaly detection, we set the number of scales $S=5$, starting from patches of size $256 \times 256$. 

Our autoencoder features a simple architecture, with an encoder $\mathcal E$ made of 5 convolutional layers each, deployed with a kernel size of 4 and stride 2, and a symmetric decoder $\mathcal D$. We construct the training set by randomly cropping 50,000 overlapping patches from normal images. We train the autoencoder for 400 epochs, using the ADAM optimizer \cite{adam} with an initial learning rate of $1 \cdot 10^{- 3}$ and a weight decay set to 0.5 every 20 epochs.

\subsection{CW-SSIM for Anomaly Detection}\label{subsec:reco_based}
We employ CW-SSIM also at inference time, to estimate the anomaly masks by comparing the original and reconstructed test images. To do so, we first divide each test image $I$ into patches, following a regular grid sampling with stride 16. Then, we reconstruct these patches through our autoencoder, and merge them to form a full-size reconstructed image $J$, averaging the overlapping regions. To identify anomalous regions in $I$, we first apply the subband decomposition on both $I$ and $J$, divide the two images into $L^\prime$ $R\times R$ windows $I^m_l$, $J^m_l$, where $m\in\{1,\ldots,M\}$ and $l\in\{1,\ldots,L^\prime\}$, and then compare these windows by the CW-SSIM index \eqref{eq:cw-ssim}. We define the anomaly score $\mathcal{S}(l)$ for each pair $I^m_l$, $J^m_l$ as the average CW-SSIM index over all the subbands: 
\begin{equation}\label{eq:cw-ssim_pool_reco}
    \mathcal{S}(l) = 1 - \frac{1}{M} \sum_{m=1}^{M} \text{CW-SSIM}(I^m_l,J^m_l).
\end{equation}
Then, we compute the anomaly score $\hat{\omega}(i,j)$ for each pixel $(i,j)$ by averaging the scores $\mathcal{S}(l)$ of the windows containing the pixel $(i,j)$. The CW-SSIM window size $R$ and number of orientations $O$ in the subband decomposition are the same used during training ($R=7$ and $O=6$), in order to extract the same type of features. To produce more robust results, we average the anomaly maps obtained computing the CW-SSIM index several times, changing the number of scales $S\in\{7,8,9\}$. Using a larger number of scales increases the dimensionality reduction factor, thus compensating for the fact that, at test time, we compare full-size images, while during training the CW-SSIM index is computed only on small patches. Then, we generate the estimated anomaly mask $\hat{\Omega}$ by detecting as anomalous those pixels $(i,j)$ whose anomaly score exceeds a threshold $\gamma$, i.e., $\hat{\omega}(i,j) \geq \gamma$. We define the threshold $\gamma$ to obtain a false positive rate (FPR) of 0.05 over the validation images, which are defective images that will be no used during test. Finally, we apply circular erosion to $\hat{\Omega}$ over a disk of radius 10 pixels to remove outliers and to reduce jagged edges, thus improving the coverage of anomalous regions, as shown in~\cite{carrera2016defect}.

\section{Experiments}\label{sec:experiments}
Here we present the experiments we performed to evaluate the anomaly detection performance of our solution. First, we describe the datasets (Section~\ref{subsec:datasets}) and the figures of merit we employ to test our solution (Section~\ref{subsec:figures}). Then we briefly present the methods against which we compare our solution (Section~\ref{subsec:methods}), and finally we illustrate and discuss our experimental results (Section~\ref{subsec:results}). 

\subsection{Benchmarking Datasets}\label{subsec:datasets}
We evaluate the performance of our methods on two anomaly detection datasets: the \textit{Nanofiber} and the \textit{MVTec Texture} datasets (see Figure \ref{fig:Nanofiber_good} for examples of normal and anomalous images). The Nanofiber dataset \cite{carrera2016defect} contains images of nanofibrous material acquired by a \emph{Scanning Electron Microscope} (SEM), with a pixel size of few tens nanometers. In normal conditions, the material presents a non-periodic continuous texture of filamentous elements with a diameter of about 10 pixels. The anomalous regions consist of agglomerates of filamentous material, and can appear in any location within the image. The annotated defects have a diameter that ranges between 20 to 350 pixels. The images are grayscale (i.e., $c=1$) with 8-bit color depth ($d=8$), and have size $696 \times 1024$ pixels.

The MVTec Texture dataset is the portion of the MVTec anomaly detection dataset~\cite{bergmann2021mvtec} that contains only textured images. We exclude images depicting objects, which are out of the scope of this paper. There are 5 different categories of either regular (\emph{Carpet}, \emph{Grid}) or random textures (\emph{Leather}, \emph{Tile}, \emph{Wood}). Test images can contain a variety of defects, such as surface anomalies (e.g., scratches, cuts), or missing parts, with a diameter between 40 to 390 pixels. The dataset is mainly composed of RGB images (i.e., $c=3$), apart from the Grid images, which are grayscale. All the images have 8-bit color depth and spatial dimension ranging between $700 \times 700$ and $1024 \times 1024$. Due to the design of CW-SSIM, we convert RGB images to grayscale before presenting them to our autoencoder.

\subsection{Figures of Merit}\label{subsec:figures}
We assess the anomaly detection performance by the area under the ROC curve (AUC), a well-known threshold-independent performance metric, computed from the anomaly scores of each pixel in test images. To enable a direct comparison with the results in \cite{bergmann2021mvtec}, in our experiments on the MVTec dataset we compute the AUC up to $\text{FPR}=0.3$ (normalized so that the maximum attainable value is 1). On the Nanofiber dataset, we also evaluate the \emph{defect coverage}, which compares the connected components of the ground-truth and the estimated anomaly mask obtained by a detector yielding $\text{FPR}=0.05$. This metric was proposed in~\cite{carrera2016defect} to assess the detection performance independently from the size of the anomalous regions, which influences the AUC since larger anomalies contain more pixels~\cite{carrera2016defect}.

\subsection{Considered Methods}\label{subsec:methods}
In our experiments we compare our solution with autoencoders sharing the same architecture but trained with different loss functions: MSE, SSIM as in~\cite{bergmann2018improving}, and MS-SSIM~\cite{MS-SSIM}. For a fair comparison, we re-train our SSIM autoencoder rather than reporting the results from~\cite{bergmann2018improving}, whose autoencoder has a simpler architecture than ours. We also evaluate the method presented in~\cite{napoletano2018anomaly}, which we indicate by \emph{Feature Dictionary}, and the Student-Teacher method~\cite{bergmann2020uninformed}, which represents the state of the art on the MVTec datasets. On the Nanofiber dataset, we also evaluate the sparse representation-based method proposed in~\cite{carrera2016defect}, which we refer to as \emph{Sparse Coding}. The SSIM and MS-SSIM autoencoders and Sparse Coding are designed to process grayscale images, thus it is necessary to convert RGB images to grayscale before applying these methods.

We report the anomaly detection performance on the Nanofiber dataset in~\cite{napoletano2018anomaly} for Feature Dictionary and in~\cite{carrera2016defect} for Sparse Coding, and the performance on the MVTec Texture datasets of Feature Dictionary and Student-Teacher as reported in~\cite{bergmann2021mvtec}. Since the Student-Teacher implementation is not publicly available, we were not able to test it on the Nanofiber dataset.

\subsection{Results and Discussion}\label{subsec:results}
In Table~\ref{tab:nano} we report the AUC and median defect coverage of the considered methods on the Nanofiber dataset~\cite{carrera2016defect}. These results show that our CW-SSIM autoencoder is the best at detecting anomalous regions, slightly outperforming the state-of-the-art Feature Dictionary~\cite{napoletano2018anomaly} in terms of AUC. The difference is even more substantial in defect coverage, meaning that our solution is superior in detecting also small anomalous regions.

\begin{table}[t]
\centering
\caption{AUC and defect coverage of the considered methods on the Nanofiber dataset. Our solution is the best method, outperforming the state of the art on this dataset.
}
\label{tab:nano}
\resizebox{\textwidth}{!}{
\begin{tabular}{|>{\centering}p{2cm}||>{\centering}p{2cm}|>{\centering}p{2cm}|>{\centering}p{2cm}|>{\centering}p{2cm}|>{\centering}p{2cm}|>{\centering}p{2cm}|}
\hline
Metrics & Sparse Coding \cite{carrera2016defect} & Feature Dict. \cite{napoletano2018anomaly} & MSE Autoencoder & SSIM Autoencoder & MS-SSIM Autoencoder & CW-SSIM Autoencoder \tabularnewline
\hline
AUC & 0.926  & 0.974 & 0.665 & 0.961 & 0.962 & \textbf{0.977}\tabularnewline
\hline
Def. Cov. & 0.650 & 0.850 & 0.000 & 0.880 & 0.870 & \textbf{0.960}\tabularnewline
\hline
\end{tabular}
}
\end{table}

\begin{table}[t!]
\centering
\caption{Normalized AUC of the considered methods of the MVTec Texture dataset. We indicate by (G) grayscale textures and by (C) color textures. Our solution outperforms the state of the art in three out of five textures from this dataset.}
\label{tab:mvtec}
\resizebox{\textwidth}{!}{
\begin{tabular}{|>{\centering}p{2cm}||>{\centering}p{2cm}|>{\centering}p{2cm}|>{\centering}p{2cm}|>{\centering}p{2cm}|>{\centering}p{2cm}|>{\centering}p{2cm}|}
\hline
Dataset & Feature Dict. \cite{napoletano2018anomaly} & Student- Teacher \cite{bergmann2020uninformed} & MSE Autoencoder & SSIM Autoencoder & MS-SSIM Autoencoder & CW-SSIM Autoencoder \tabularnewline
\hline
Carpet (C) & 0.943 & 0.927 & 0.543 & 0.754 & 0.753 & \textbf{0.947}\tabularnewline
\hline
Grid (G) & 0.872 & 0.974 & 0.662  & 0.852 & 0.927 & \textbf{0.978}\tabularnewline
\hline
Leather (C) & 0.819 & 0.976 & 0.658 & 0.800 & 0.696 & \textbf{0.979}\tabularnewline
\hline
Tile (C)  & 0.854 & \textbf{0.946} & 0.552 & 0.650 & 0.679 & 0.847\tabularnewline
\hline
Wood (C) & 0.720 & \textbf{0.895} & 0.574 & 0.706 & 0.640 & 0.753\tabularnewline
\hline
\end{tabular}}
\end{table}

Table~\ref{tab:mvtec} reports the normalized AUC (up to $\text{FPR}=0.3$) of the considered methods on the different categories in the MVTec Texture dataset~\cite{bergmann2021mvtec}. Our CW-SSIM autoencoder represents the best-performing method on three out of five categories of textures, namely Grid (which is grayscale), Carpet and Leather (which are color). We remark that our solution, which is based on a simple architecture with $5.57\cdot10^6$ trainable parameters, outperforms the state-of-the-art Feature Dictionary and Student-Teacher approaches, which leverage comparably more complicated pre-trained models with respectively $11.46\cdot10^6$ and $26.07\cdot10^6$ parameters. Most remarkably, our solution yields better AUC than competing methods also on color textures like Carpet and Leather, even though we convert RGB images to grayscale before feeding them to our autoencoder, while Feature Dictionary and Student-Teacher take as input the original RGB images. In contrast, on Tile and Wood, our solution performs similarly to Feature Dictionary but worse than Student-Teacher. We speculate that this is due to the fact that color information is crucial to detect anomalies in these textures, and this motivates us to further investigate how to manage colors in our solution.

\begin{figure}[t]
\includegraphics[width=\textwidth]{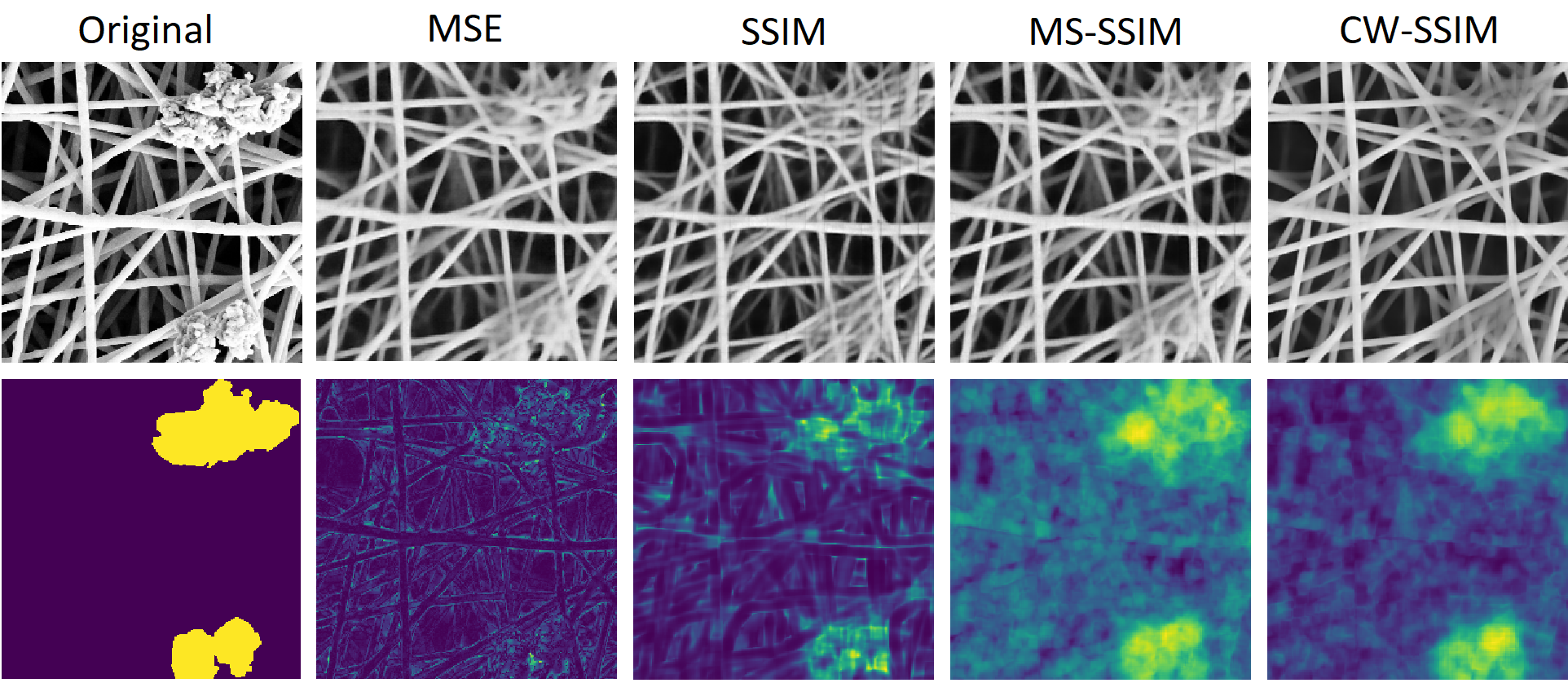}
\centering
\caption{An example of reconstruction and anomaly scores produced by autoencoders trained with different loss functions from a Nanofiber image. This image shows that autoencoders trained with structural similarity metrics, and CW-SSIM in particular, yield better reconstruction quality and superior anomaly detection performance than a traditional MSE autoencoder.}\label{fig:reconstruct}
\end{figure}

The results in Tables~\ref{tab:nano} and~\ref{tab:mvtec} confirm that a loss function based on SSIM substantially improve the anomaly detection performance of autoencoders compared to using the MSE, as shown in~\cite{bergmann2018improving}. Our experiments show that using MS-SSIM yields similar performance to SSIM, while our loss function and anomaly detection procedure based on CW-SSIM further improve the results, making our solution comparable or superior to the state of the art. This is due to the fact that autoencoders based on structural similarity metrics, and CW-SSIM in particular, achieve higher reconstruction quality of normal images than MSE autoencoders, and this yields superior anomaly detection performance. We qualitatively illustrate this in Figure~\ref{fig:reconstruct}, which shows an image from the Nanofiber dataset, its reconstruction produced by autoencoders based on MSE, SSIM, MS-SSIM, and CW-SSIM, and the corresponding anomaly maps $\hat\omega$ compared to the ground truth.

\section{Conclusions and Future Work} \label{sec:conclusions}
We propose an effective training strategy for autoencoders to detect anomalous regions in textured images. In particular, we employ a custom loss function based on CW-SSIM~\cite{wang2005translation} for both training and testing. Our experiments show that our solution improves the state of the art on several anomaly detection benchmarks, outperforming deep neural networks with more sophisticated and computationally demanding architectures. Although CW-SSIM is designed for grayscale images, thus it cannot exploit color information, our solution can effectively detect anomalous regions also from RGB images converted to grayscale. Future work will investigate how to combine CW-SSIM and \textit{Optimal Color Composition Distance} (OCCD)~\cite{mojsilovic2002extraction} in our custom loss function for autoencoders, as suggested in~\cite{zujovic2009structural}, since in some cases color information might be crucial for anomaly detection. Moreover, we will make our autoencoder fully convolutional to achieve more efficient training and testing.

\section*{Acknowledgements}
We gratefully acknowledge the support of NVIDIA Corporation with the four RTX A6000 GPUs granted through the Applied Research Accelerator Program to Politecnico di Milano.

%
%
\bibliographystyle{splncs04}
\bibliography{mybibliography}
\end{document}